\documentclass[10pt,twocolumn,letterpaper]{article}

\usepackage{wacv}
\usepackage{times}
\usepackage{epsfig}
\usepackage{graphicx}
\usepackage{amsmath,amssymb} 
\usepackage{color}
\usepackage{cite}
\usepackage{multirow}
\usepackage{subcaption}
\usepackage{setspace}

\usepackage[pagebackref=true,breaklinks=true,letterpaper=true,colorlinks,bookmarks=false]{hyperref}

\graphicspath{{fig/}}
\DeclareMathOperator*{\argmin}{arg\,min}

\wacvfinalcopy 

\def\wacvPaperID{20} 

\ifwacvfinal\fi
\begin{document}

\title{Looking at Outfit to Parse Clothing}

\author{
	Pongsate Tangseng \\
	Tohoku University\\
	{\tt\small tangseng@vision.is.tohoku.ac.jp}
	\and
	Zhipeng Wu \\
	The University of Tokyo\\
	{\tt\small zhipeng\_wu@ipc.i.u-tokyo.ac.jp}
	\and
	Kota Yamaguchi \\
	Tohoku University\\
	{\tt\small kyamagu@vision.is.tohoku.ac.jp}
}



\maketitle
\ifwacvfinal\thispagestyle{empty}\fi

\begin{abstract}
This paper extends fully-convolutional neural networks (FCN) for the clothing parsing problem. Clothing parsing requires higher-level knowledge on clothing semantics and contextual cues to disambiguate fine-grained categories. We extend FCN architecture with a side-branch network which we refer outfit encoder to predict a consistent set of clothing labels to encourage combinatorial preference, and with conditional random field (CRF) to explicitly consider coherent label assignment to the given image. The empirical results using Fashionista and CFPD datasets show that our model achieves state-of-the-art performance in clothing parsing, without additional supervision during training. We also study the qualitative influence of annotation on the current clothing parsing benchmarks, with our Web-based tool for multi-scale pixel-wise annotation and manual refinement effort to the Fashionista dataset. Finally, we show that the image representation of the outfit encoder is useful for dress-up image retrieval application.
\end{abstract}

\section{Introduction}
Clothing parsing is a specific form of semantic segmentation, where the categories are one of the clothing items, such as {\it t-shirt}. Clothing parsing has been actively studied in the vision community~\cite{fashionCVPR12,paperdoll,CRFClothesParsing,CCP,CFPD}, perhaps because of its unique and challenging problem setting, and also because of its tremendous value in the real-world application. Clothing is an essential part of our culture and life, and a significant research progress has been made with a specific application scenario in mind~\cite{liu2012street,6618404,6595844,GettingTheLook,7153120,7045985,Wang_et_al,Kiapour_2015_ICCV,Veit_2015_ICCV,Liu_2016_CVPR,Liu2016}.
In this paper, we consider how we can utilize recent deep segmentation models in clothing parsing and discuss issues in the current benchmarks.

\begin{figure}[t]
  \centering
    \begin{subfigure}{0.5\columnwidth}
        \includegraphics[width=\linewidth]{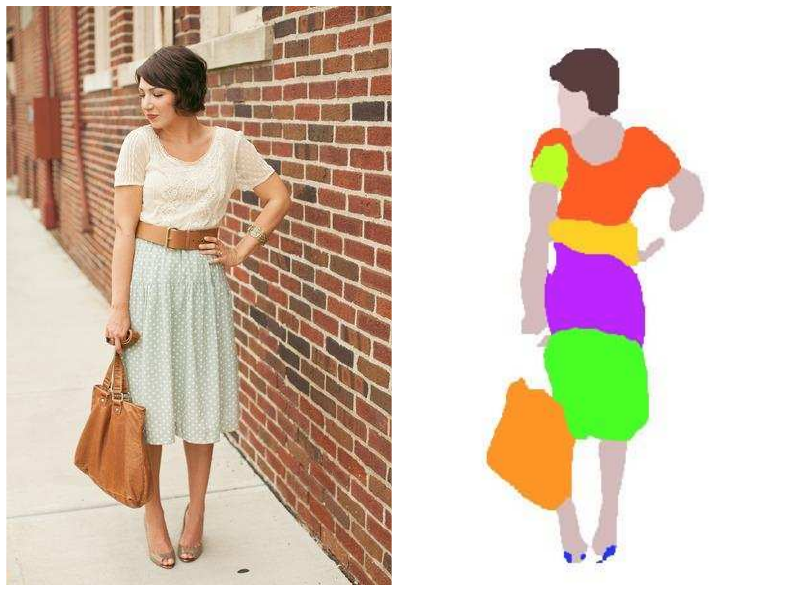}
    \end{subfigure}
    \begin{subfigure}{0.15\columnwidth}
        \begin{scriptsize}
          \parbox{\columnwidth}{
			\raisebox{1ex}{\fcolorbox[RGB]{0,0,0}{255,67,0}{}} T-shirt \\
            \raisebox{1ex}{\fcolorbox[RGB]{0,0,0}{255,133,0}{}} bag \\
            \raisebox{1ex}{\fcolorbox[RGB]{0,0,0}{255,200,0}{}} belt \\
            \raisebox{1ex}{\fcolorbox[RGB]{0,0,0}{177,255,0}{}} blouse \\
            \raisebox{1ex}{\fcolorbox[RGB]{0,0,0}{44,255,0}{}} dress \\
          }
        \end{scriptsize}
    \end{subfigure}
    \begin{subfigure}{0.15\columnwidth}
        \begin{scriptsize}
          \parbox{\columnwidth}{
            \raisebox{1ex}{\fcolorbox[RGB]{0,0,0}{226,196,196}{}} face \\
            \raisebox{1ex}{\fcolorbox[RGB]{0,0,0}{64,32,32}{}} hair \\
            \raisebox{1ex}{\fcolorbox[RGB]{0,0,0}{0,22,255}{}} shoe \\
            \raisebox{1ex}{\fcolorbox[RGB]{0,0,0}{206,176,176}{}} skin \\
            \raisebox{1ex}{\fcolorbox[RGB]{0,0,0}{177,0,255}{}} skirt \\
          }
        \end{scriptsize}
    \end{subfigure}
    \\
    \begin{subfigure}{.7\columnwidth}
    	\includegraphics[width=\linewidth]{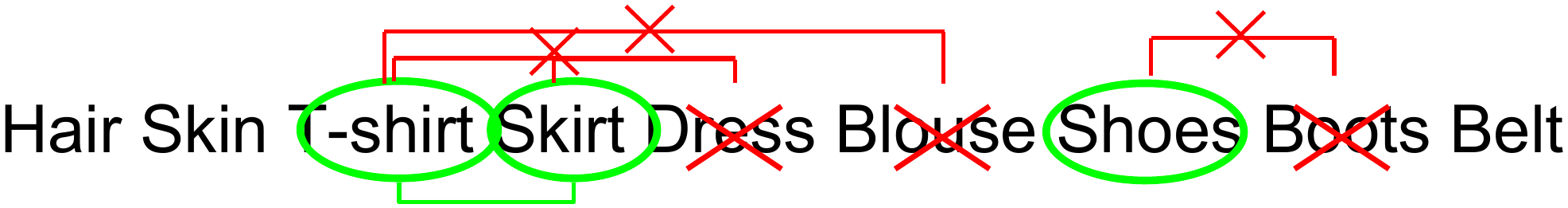}
    \end{subfigure}
    
    \caption{Combinatorial preference in clothing parsing: dress and skirt are in the exclusive relationship, yet independent pixel-wise prediction cannot encode such knowledge and results in mixture of patches (FCN-8s~\cite{FCN}). We propose the side-path outfit encoder and CRF alongside the segmentation pipeline to address the issue.
    }
    \label{1stfig}
\end{figure}

Clothing parsing distinguishes itself from general object or scene segmentation problems in that fine-grained clothing categories require higher-level judgment based on the semantics of clothing and the deforming structure within an image. What we refer the semantics here is the specific type of clothing combination people choose to wear in daily life. For example, people might wear {\it dress} or separate {\it top} and {\it skirt}, but not both of them together. However, from recognition point of view, both styles can look locally very similar and can result in false positives in segmentation, as shown in Figure~\ref{1stfig}. Such combinatorial preference at the semantics level~\cite{McAuley:2015:IRS:2766462.2767755,yamaguchi2015mix,Veit_2015_ICCV} introduces a unique challenge in clothing parsing where a bottom-up approach is insufficient to solve the problem~\cite{Dong2014}.

In this paper, we approach the clothing parsing problem using fully-convolutional neural networks (FCN). FCN has been proposed for general object segmentation~\cite{FCN} and shown impressive performance thanks to the rich representational ability of deep neural networks learned from a huge amount of data. To utilize FCN in clothing parsing, we need to take the above clothing-specific challenges into account, as well as a care to address the lack of training data for learning large neural networks. Based on the FCN architecture, we propose to extend the parsing pipeline by 1) a side-branch that we call \emph{outfit encoder} to predict the combinatorial preference of garments for dealing with semantics-level consistency, and 2) conditional random field (CRF) to consider both semantics and appearance-level context in the prediction. Experimental results show that, starting from a pre-trained network, we are able to learn the outfit encoder and fine-tune the whole segmentation network with a limited amount of training data, and our model achieves the state-of-the-art performance in the publicly available Fashionista dataset~\cite{fashionCVPR12} and Colorful Fashion Parsing dataset (CFPD)~\cite{CFPD}.

We also study the qualitative issue in the current clothing datasets. The existing benchmarks suffer from erroneous annotations due to the limitation in the superpixel-based annotation, as well as from the ambiguity of labels~\cite{CRFClothesParsing}. We develop a Web-based tool to interactively annotate pixels at multiple scales, and study how much influence we have on the performance metrics by manually refining the Fashionista dataset~\cite{fashionCVPR12}.

The outfit encoder learns a compact representation of the combinatorial clothing preference through the training of segmentation pipeline. We find that the resulting internal representation of the encoder is suitable for style retrieval application. The learned representation compactly encodes the gist of the dress-up style of the picture, and when used in retrieval, the representation is able to find semantically similar clothing style (e.g., dress only or shirt + skirt combination), even if the low-level appearance cues such as color or texture look different.

We summarize our contribution in the following.
\begin{itemize}
  \item We propose the side-path outfit encoder for the FCN architecture, and together with CRF to improve segmentation performance in clothing parsing. The evaluation shows that our model achieves state-of-the-art performance in clothing parsing.
  \item We develop a Web-based tool to interactively annotate pixels and study the qualitative influence in the segmentation benchmarks. Using the tool, we manually annotate the Fashionista benchmark~\cite{fashionCVPR12} with high-quality and less ambiguous labels, which we refer Refined Fashionista dataset, and study the impact on the benchmark. We will release the tool and annotation for future research.
  \item We make a preliminary study showing that the outfit encoder is also useful for retrieval. The encoder representation compactly encodes the combinatorial preference of clothing items, and suitable for retrieving semantically similar styles.
\end{itemize}

\section{Related Work}\label{sec:related_work}

\subsection{Semantic segmentation}
The use of deep convolutional neural networks for semantic segmentation is increasingly becoming popular since the recent success in dense object recognition~\cite{sermanet2013overfeat,FCN,BansalChen16}. Various techniques have been proposed to further improve the performance of dense prediction by deep neural networks, including global context information~\cite{co-cnn,gdn}, learning deconvolution layers~\cite{deconvnet}, applying conditional random fields as a post-processing~\cite{deeplab,crfasrnn}, or incorporating weakly annotated data in training set~\cite{weaklyandsemi}. In this paper, we propose a side-path encoder to predict unique set of consistent labels in segmentation and feed FCN output to fully-connected CRF for addressing combinatorial preference issue in clothing parsing. The side-path can be considered one kind of attention mechanism~\cite{Xu2015show} or gating function to control information flow~\cite{DBLP:journals/corr/ChoMGBSB14}.

\begin{figure*}[t]
  \centering
  \includegraphics[width=.75\textwidth]{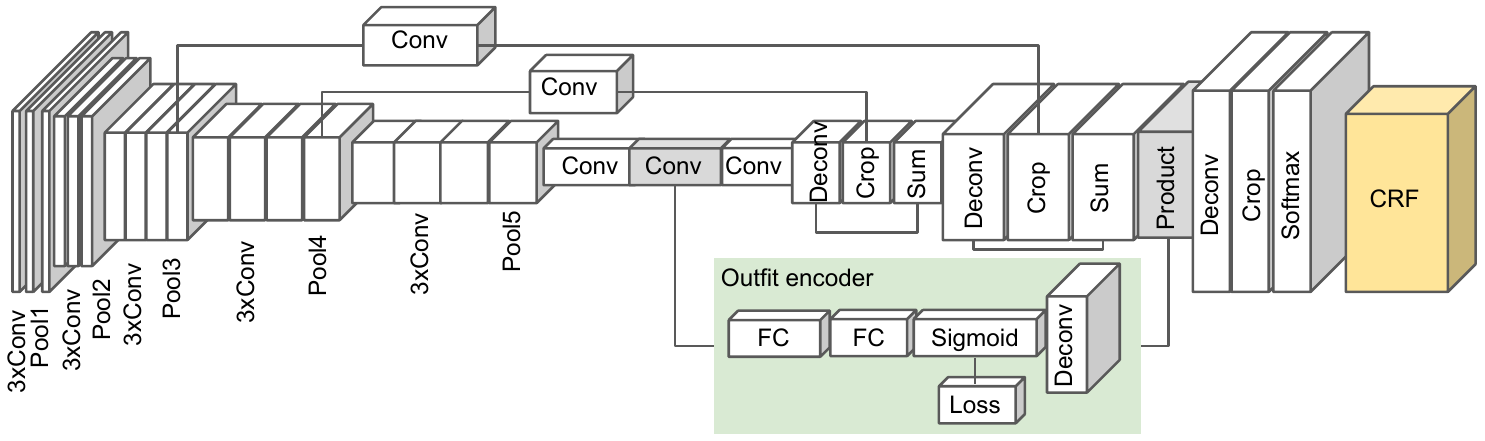}
  \caption{Our segmentation model based on FCN. We introduce 1) the outfit encoder to filter inappropriate clothing combination from segmentation, and 2) CRF to assign a visually consistent set of clothing labels.}
  \label{fig:network}
\end{figure*}

\subsection{Clothing parsing}
Clothing parsing has been an active subject of research in the vision community~\cite{fashionCVPR12,crfunrestrict,Yamaguchi2015,CRFClothesParsing,CCP,CFPD}. Also, there is a similar variant of parsing problem referred human parsing~\cite{Dong2013,Dong2014,ATR,Liu2015,Liu2015matching,co-cnn}. The major difference between clothing and human parsing are the types of labels; Clothing parsing attempts to identify fine-grained categories of clothing items such as {\it t-shirt} and {\it blouse}, whereas human parsing aims at identifying body parts and broad clothing categories, such as {\it left-leg} or {\it upper-body clothing}. 
This difference brings a further challenge in clothing parsing that we have to disambiguate confusing labels, for example, {\it sweater} and {\it top}, even if they look similar. 

The basic approach attempted in clothing parsing is first to identify human body configuration in the image, and given the body joints, solve for the best assignment of pixel-wise labels~\cite{fashionCVPR12,crfunrestrict,CRFClothesParsing,Yamaguchi2015,Liu2015}. Often superpixel-based formulation is employed instead of pixel-wise labeling to reduce the computational complexity, though superpixels sometimes harm the final segmentation quality in the presence of textured region~\cite{CRFClothesParsing}. In this work, we rely on the large size of receptive fields in the deep architecture to identify the contextual information from human body, and eliminate the extra pre-processing necessity to explicitly identify human parts in the image. Skipping pose-estimation has an additional advantage of not requiring full-body visible in the image frame.


\subsection{Clothing retrieval}
Retrieval and recommendation is one of the most important applications in clothing recognition, and there have been many efforts in various scenario, such as street-to-shop~\cite{liu2012street,Kiapour_2015_ICCV}, or style suggestion or matching~\cite{GettingTheLook,Veit_2015_ICCV}. The key idea is to learn a meaningful representation to define the distance between style images~\cite{7045985,Simo-Serra_2016_CVPR}. In this paper, we consider retrieving the whole dress-up style rather than looking at a specific item using the outfit encoder representation.


\section{Our approach} \label{attr_branch}

This section describes our FCN model with outfit encoder and fully-connected CRF.

\subsection{Fully convolutional networks} \label{sec:fcn}

We build our segmentation model upon the FCN architecture~\cite{FCN}. The FCN model is a convolutional neural network and makes a dense pixel-wise prediction by replacing all fully-connected layers in the classification network with convolutional layers, followed by upsampling filters to recover the original image resolution in the output. The FCN model proposed in \cite{FCN} can implement a different upsampling strategy. In this paper, we use the 8-stride variant of the VGG 16-layer net \cite{VGG} to build our clothing parsing model on.

We choose the FCN architecture for clothing parsing expecting that the receptive fields of mid-to-later layers can cover sufficiently large input regions so that the final pixel-wise prediction makes a proper judgment on different but similar items, such as {\it coat} and {\it jacket}, based on the contextual information in the image.
Thanks to the deep architecture, the receptive field in the last layer has large coverage of the input frame and is expected to contain sufficient contextual information from human body in the internal representation within the network. Also, there is no restriction on the input image that the image frame must contain the full human body.

\subsection{Outfit encoder and filtering} \label{sec:sege_attrlog}

We introduce a side path to FCN that encode and predict combinatorial preference of garment items. Figure \ref{fig:network} illustrates the architecture of our network. The idea is to predict the \emph{garment combination as a summary of segmentation} using this side encoder path, and feed it into the main segmentation pipeline to filter the prediction on the possible set of labels. Our outfit encoder predicts a binary indicator of existence of each clothing label in the final segmentation, as usual attribute prediction problem.

The outfit encoder inserts two fully-connected (FC) layers and a sigmoid layer to predict a vector of clothing indicators. The first FC layer has 256 dimensions, and the second FC layer has dimensions equal to the number of classes in the dataset. The second layer predicts confidences of existence of each garment, which can be viewed as soft-attention or gating function to the segmentation pipeline. We connect 2nd FC with a sigmoid, then merge this vector back to the FCN segmentation pipeline using element-wise product, similarly to gating functions in LSTM or GRN~\cite{DBLP:journals/corr/ChoMGBSB14}. This vector multiplies confidences to filter out uncertain labels from the image.

Formally, our outfit filtering is expressed by the following. Let us denote the heat-maps of the FCN by $F_i$ for each label $i$, and the scalar prediction by our outfit encoder by $g_i$. Then, we apply a product to obtain the filtered heat-maps $G_i$:
\[
G_i = g_i \cdot F_i.
\label{eqn:filtering}
\]

The role of outfit filtering is to encourage or prevent certain clothing combination from appearing at the image-level (e.g., skirt + dress never happens together, but dress-only or skirt + top likely). Such decision requires to look at the whole image instead of local parts, and thus we let the garment encoder predict the existence of all labels at the image-level. The prediction is integrated back to the main segmentation trunk as a bias to the heat-maps produced per label.

The internal representation of the garment encoder makes a compact representation of clothing semantics in the given picture. We show that the representation is also useful in image retrieval scenario in Sec~\ref{sec:img_retrieval}.

\subsection{Conditional random fields}

The outfit filtering enforces combinatorial semantics in the segmentation, but still the prediction contains a lot of small regions of incompatible items due to the pixel-wise prediction without explicit modeling of label combinations. Here, we introduce a fully-connected CRF to improve the segmentation quality. CRF has been shown to be effective for segmentation~\cite{efficientCRF,CRFClothesParsing,densecut} or used as post-processing step after segmentation using CNN~\cite{crfasrnn,deeplab,weaklyandsemi}. In this paper, we use the implementation of \cite{efficientCRF} as a post-processing step to correct predictions\footnote{We also attempted the end-to-end model~\cite{crfasrnn} but could not reliably learn the network perhaps due to the small data size in our experiment.}.

Our energy is a fully-connected pairwise function:
\begin{equation}
    E(\pmb{x}) = \sum_i \phi (x_i) + \sum_{i<j} \psi (x_i,x_j),
\end{equation}
where \(\pmb{x}\) is the label assignment for pixels. The unary potential $\phi (x_i) = -\log{P(x_i)}$ takes label assignment probability distribution at pixel $i$, denoted by \(P(x_i)\). We use the softmax output of FCN for this probability. The pairwise potential \(\psi (x_i,x_j)\) considers contrast and position of two pixels using Gaussian kernels:

\medmuskip=-1mu
\thinmuskip=-1mu
\thickmuskip=-1mu

\begin{eqnarray}
    \psi_p(x_i,x_j) &=&
        \begin{cases}
            w_1g_1(i,j) + w_2g_2(i,j)   &\quad \mathrm{if}~x_i \neq x_j \\
        0                            &\quad \mathrm{otherwise},
        \end{cases} \\
    g_1(i,j) &=& \exp\left(-\frac{|p_i-p_j|^2}{\sigma^2_\text{position}}-\frac{|I_i-I_j|^2}{\sigma^2_\text{color}}\right), \\
    g_2(i,j) &=& \exp\left(-\frac{|p_i-p_j|^2}{\sigma^2_\text{smooth}}\right),
\end{eqnarray}
where \(g_1(i,j)\) is an appearance kernel, \(g_2(i,j)\) is a smoothness kernel, \(p_i\) is the position of pixel $i$, and \(I_i\) is the RGB color vector of pixel $i$. Weights of both kernels are controlled by \(w_1\) and \(w_2\). The hyper-parameters \(\sigma_\text{position}\), \(\sigma_\text{color}\), \(\sigma_\text{smooth}\) control position scaling, color scaling, and smoothness scaling, respectively. We find the best hyper-parameters by non-linear optimization (L-BFGS) using the validation set, starting from the initial parameters $(w_1, w_2, \sigma_\text{position}, \sigma_\text{color}, \sigma_\text{smooth}) = (10, 10, 30, 10, 3)$ in our experiment. We approximately solve for the optimal label assignment using the algorithm of \cite{efficientCRF}:
\begin{equation}
  \pmb{x}^* = \argmin _{\pmb{x}} E(\pmb{x}).
\end{equation}

\subsection{Training the network} \label{sec:training}

Our assumption is that we do not have a sufficient number of images to learn deep networks from scratch in clothing parsing. We consider transfer-learning from the pre-trained FCN models. In this paper, we follow the incremental procedure to fine-tune a coarse FCN model~\cite{FCN}, and learn our garment encoder after that. We train FCN-32s, FCN-16s, and FCN-8s in sequence, move on to the training of the encoder, then fine-tune the whole model at the end.

We need to train new layers in the outfit encoder from scratch, using binary attribute vector as a ground truth. We can trivially obtain a binary vector from segmentation ground truth by simply taking a unique set of labels. We use sigmoid cross-entropy loss to train the encoder first (Figure \ref{fig:network}). The encoder path is first trained independently with the binary indicators fixing learning rate for the main segmentation pipeline to zero, then fine-tuned together with segmentation pipeline later to avoid local optima. We implement the neural network using Caffe framework~\cite{Caffe}.


\section{Interactive pixel annotation} \label{sec:datasets}

\begin{figure}[t]
  \centering
  \includegraphics[width=\columnwidth]{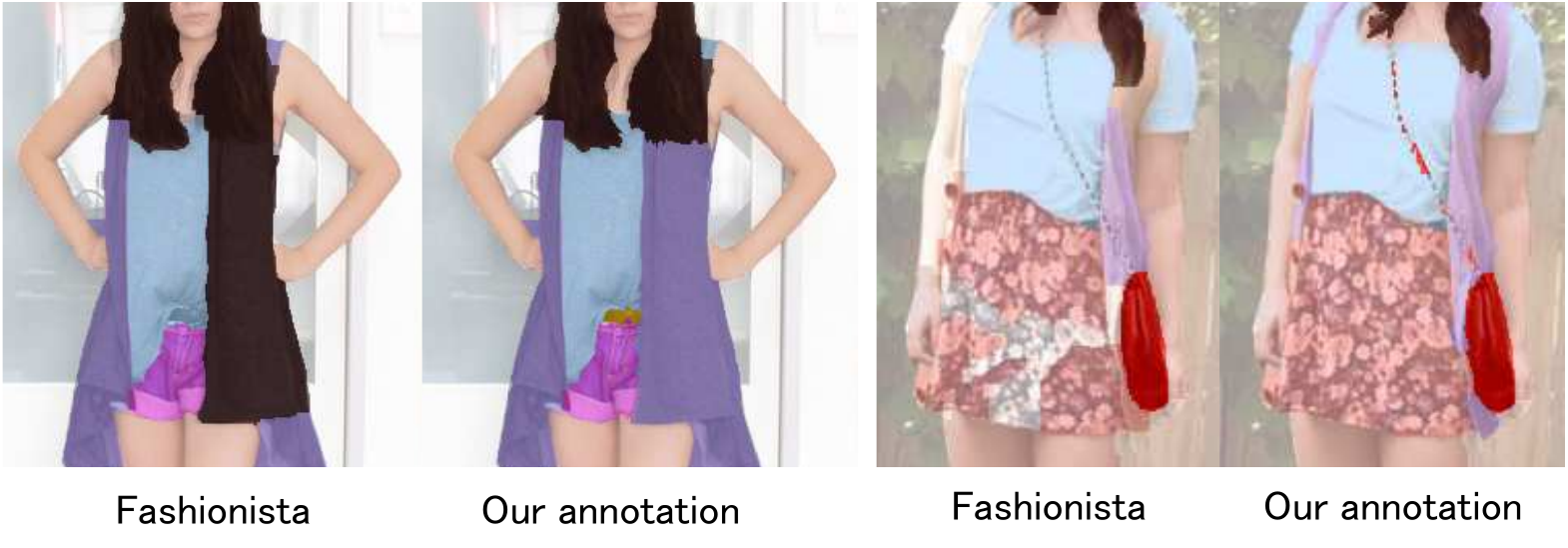}
  \caption{Common annotation error due to inappropriate superpixels. Superpixels can segment object boundary when the region is uniform and distinct from others, but spill over the object boundary when the regions have similar color (left) or produce a lot of small segments on the textured surface (right).}
  \label{fig:annotation-issues}
\end{figure}

We initially attempted to evaluate our model on the publicly available Fashionista dataset~\cite{fashionCVPR12} and CFPD~\cite{CFPD}. However, we find that both datasets have quite a bit of annotation errors. The annotation errors in both datasets are caused by 1) superpixel errors (Figure \ref{fig:annotation-issues}) and 2) ambiguous clothing categories (e.g., {\it shirt} and {\it blouse}). These annotation errors lead us to noticeable prediction errors in the final segmentation. Therefore, we decided to manually improve the annotation quality of the Fashionista dataset, and study how much quality improvement we can benefit from the dataset itself.

\begin{figure}[t]
  \centering
  \includegraphics[width=\columnwidth]{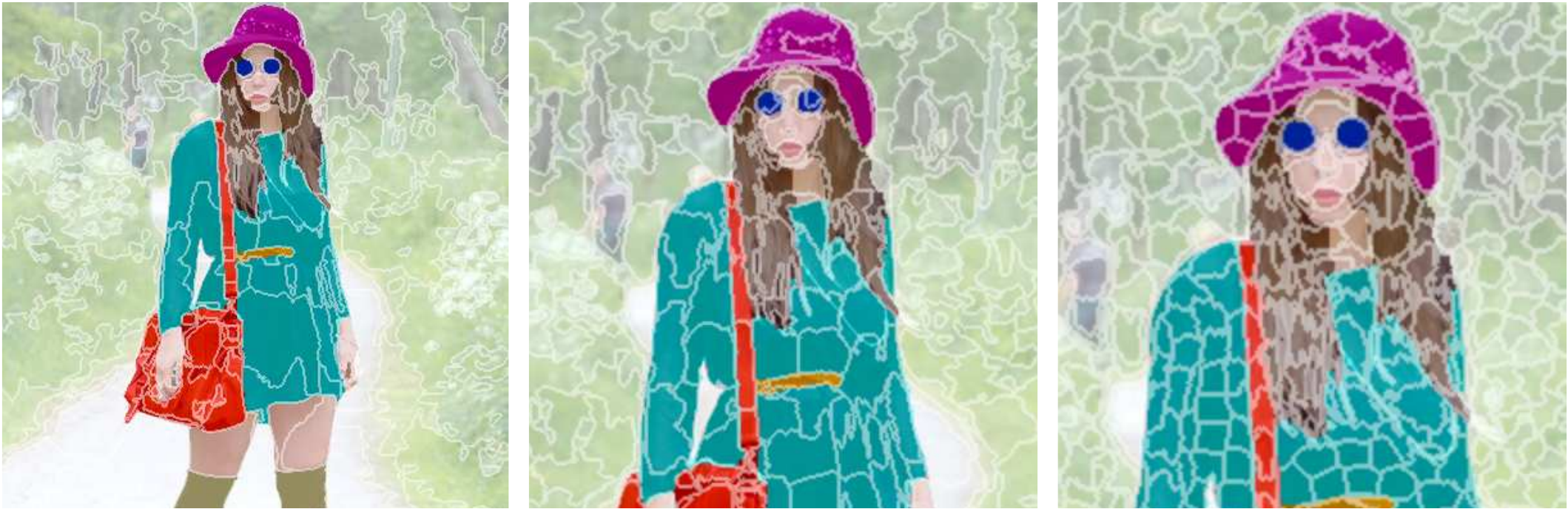}
  \caption{Coarse-to-fine interactive annotation. We compute SLIC~\cite{achanta2012slic} superpixels on the fly so that the annotator can select the desired scale.}
  \label{fig:coarse-to-fine}
\end{figure}

We have developed a Web-based tool to interactively annotate pixels. Our segmentation tool is based on annotation over superpixels, but we resolve the improper boundary of superpixels by coarse-to-fine interactive segmentation. Figure \ref{fig:coarse-to-fine} demonstrates the example. Our tool computes SLIC~\cite{achanta2012slic} superpixels on the fly inside the Web browser, and the annotator can adjust the resolution of the superpixels as needed to mark smaller segments. Our tool can overcome the limitation of superpixel-based annotation because the tool does not share the segment boundary across multiple-resolutions. The tool can also apply morphology-based smoothing to remove the artifacts of SLIC superpixels. In a modern Web browser, the tool computes SLIC superpixels in a second, depending on the size of the image.

We merged some of the confusing clothing in the 56 categories (e.g., shirt and blouse) in Fashionista dataset as well as split broad labels (e.g., accessories) to prevent ambiguity in the annotation. We manually annotated all the 685 images in the Fashionista dataset with 25 categories. Note that the mapping of the labels is not unique and unfortunately the labels in one dataset cannot be automatically converted to the other.

We hope to expand the number of fully-annotated images in the future, but we find it still challenging to scale up the dataset due to the required level of expertise for a non-expert user in crowdsourcing service. It is our future work to make our annotation tool as easy as possible for non-expert users so that any type of semantic segmentation can benefit. We release our interactive annotation tool to the community as open-source software, as well as our annotation to Fashionista dataset.

\section{Experiments} \label{exp_res}

\subsection{Datasets}
We use Fashionista with the original annotation of 56 categories~\cite{fashionCVPR12} (Fashionista v0.2), our refined annotation with 25 categories (Refined Fashionista), and CFPD~\cite{CFPD} with 23 categories. Fashionista consists of 685 front-facing full-body images. Every pixel is given one of 56 fine-grained categories. However, the dataset has only 685 annotated images and some label appears only once or twice in the dataset. This results in some skewed performance metric due to the missing category in the ground truth in the test split. Our Refined-Fashionista reduces the number of clothing categories from 56 to 25 essential labels so as to avoid ambiguous labels. The annotation contains almost no superpixel artifact. CFPD consists of 2,682 annotated images based on superpixels for 23 labels. We divide Fashionista dataset into train-test splits using the same setting to the previous work~\cite{fashionCVPR12}, with 10\% of training images leaving for validation, and divide CFPD dataset into (train, validation, test) = (78\%, 2\%, 20\%) ratio. Images in all datasets are $400\times600$ pixels in RGB color, and we do not change the image format in our experiments. Each image shows a front-facing person with visible full-body.

\subsection{Evaluation methods}
We measure the performance using pixel-wise accuracy and intersection-over-union (IoU). We compare our models against FCN-32s, FCN-16s and FCN-8s~\cite{FCN}, as well as against the reported state-of-the-art for each dataset, though some measurements are not available in the respective publication and the experimental condition could be slightly different. Note that we cannot directly compare the performance against human parsing~\cite{co-cnn} due to the difference in semantic categories and the difficulty of reproducing the same condition without a publicly available dataset.

\subsection{Quantitative evaluation}
Table \ref{tableCompare} shows performance of our models compared to various baselines. First, we notice that using FCN already shows the solid performance improvement over the previous state-of-the-art~\cite{Yamaguchi2015,CRFClothesParsing,CFPD} that are not based on deep architecture. Also, as reported in~\cite{FCN}, applying finer-scale upsampling (8s) improves the segmentation quality. Our best model achieves the state-of-the-art 88.68\% accuracy compared to 84.68\% of \cite{Yamaguchi2015} and 54.65\% IoU compared to 42.10\% of \cite{CFPD}, even with annotation issues in both benchmarks. In our Refined Fashionista dataset, our model marks 51.78\% of IoU, which is a significant improvement from 44.72\% of the base FCN-8s model.

Our outfit filtering with CRF makes an improvement in all of the datasets. Particularly, CRF shows an improvement in all cases in all datasets compared to the model without CRF. This confirms the lack of joint prediction ability in the plain FCN~\cite{deeplab,crfasrnn}. The final effect of outfit filtering varies depending on the dataset. In Fashionista v0.2, our outfit filtering has weak effect, and only CRF is showing a noticeable improvement. We suspect this is due to the large number of ambiguous categories in Fashionista v0.2, such as {\it blazer} and {\it jacket}. Outfit filtering has weak influence by itself in CFPD, but combined with CRF, achieves the best IoU. In Refined Fashionista, our outfit encoder together with CRF achieves the best performance. We suspect the difference between datasets partly stems the low-quality annotation in Fashionista v0.2 and CFPD.

\begin{table}[t]
  \centering
  \caption{Parsing performance [\%].}
  \scriptsize
  \label{tableCompare}
  \resizebox{\columnwidth}{!}{
  \begin{tabular}{|l|l|c|c|}
  \hline
  Dataset                           & Method              & Acc   & IoU   \\ \hline
  \multirow{8}{*}{\shortstack[l]{Fashionista\\v0.2}}
                                    & Paper doll \cite{Yamaguchi2015}  & 84.68 & -   \\ 
                                    & Clothelets CRF \cite{CRFClothesParsing}   & 84.88 & - \\
                                    & FCN-32s \cite{FCN}   & 85.94 & 29.61 \\
                                    & FCN-16s \cite{FCN}      & 87.53 & 34.26   \\
                                    & FCN-8s \cite{FCN}   & 87.51 & 33.97 \\ 
                                    & \hspace{.5ex} + CRF       & \bf 88.68 & \bf 38.03 \\ 
                                    & \hspace{.5ex} + Outfit filter & 87.55 & 34.26 \\ 
                                    & \hspace{.5ex} + Outfit filter + CRF  & 88.34 & 37.23 \\ \hline
  \multirow{6}{*}{\shortstack[l]{Refined\\Fashionista}}
                                    & FCN-32s \cite{FCN}   & 88.56 & 40.88 \\
                                    & FCN-16s \cite{FCN}      & 89.74 & 43.96   \\
                                    & FCN-8s \cite{FCN}   & 90.09 & 44.72 \\ 
                                    & \hspace{.5ex} + CRF       & 91.23 & 49.21 \\ 
                                    & \hspace{.5ex} + Outfit filter        & 91.50 & 46.40 \\ 
                                    & \hspace{.5ex} + Outfit filter + CRF      & \bf 91.74 & \bf 51.78 \\ \hline
  \multirow{7}{*}{CFPD}             & CFPD \cite{CFPD}    & -   & 42.10  \\ 
                                    & FCN-32s \cite{FCN}   & 90.34  & 47.65  \\
                                    & FCN-16s \cite{FCN}      & 91.27 & 50.07   \\
                                    & FCN-8s \cite{FCN}   & 91.58 & 51.28 \\ 
                                    & \hspace{.5ex} + CRF       & \bf 92.39 & 54.60 \\ 
                                    & \hspace{.5ex} + Outfit filter & 91.52 & 51.42 \\ 
                                    & \hspace{.5ex} + Outfit filter + CRF & 92.35 & \bf 54.65 \\ \hline
  \end{tabular}}
\end{table}

\begin{figure*}[t]
  \centering
  \includegraphics[width=\textwidth]{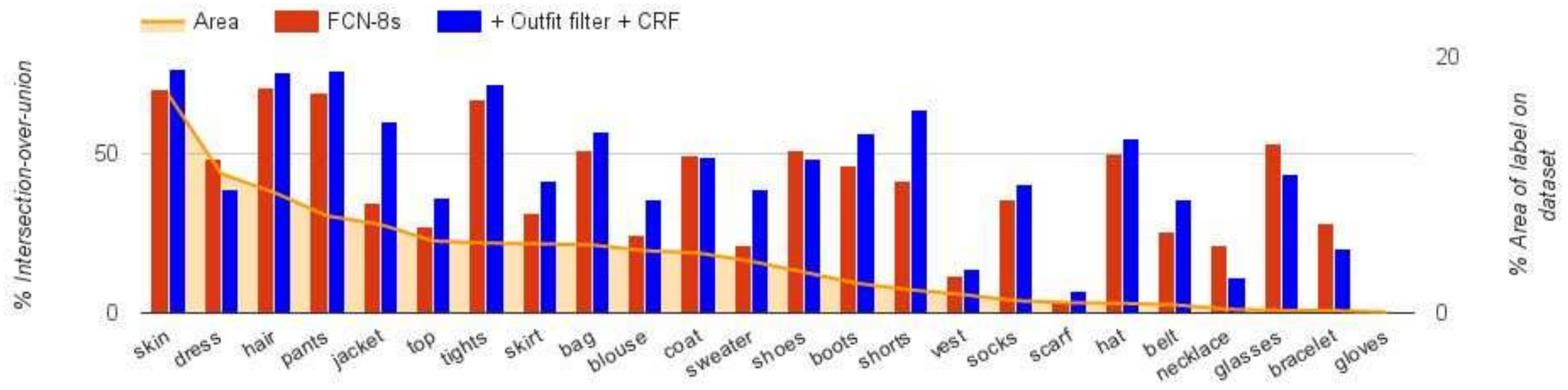}
  \caption{Intersection-over-union (IoU) in Refined Fashionista dataset [\%].}
  \label{fig:iou}
\end{figure*}

Figure \ref{fig:iou} plots the Intersection-over-union (IoU) of FCN-8s and our model in our Refined Fashionista dataset, with \% area of each class in the dataset. We find that our models improve IoU in almost all categories. The exceptions are small items, such as {\it necklace}, {\it glasses}, or {\it bracelet}, and {\it dress}. Dresses are usually confused with {\it blouse} and {\it skirt} combination. The small items tend to be smoothed out by CRF, and does not always yield precise segmentation. We also find that it is challenging even for human to give precise annotation on small items, such as bag straps or necklace. Unfortunately, our annotation tool in Sec \ref{sec:datasets} is unable to provide annotation on such small regions, but we have less practical importance in identifying almost invisible items anyway. All models get 0 IoU for {\it gloves} due to the extremely limited examples (10 out of 685 images) in the Fashionista dataset.

\subsection{Qualitative evaluation}\label{sec:qualitative}

\begin{figure}[t]
  \centering
    \begin{subfigure}{\columnwidth}
        \includegraphics[width=\linewidth]{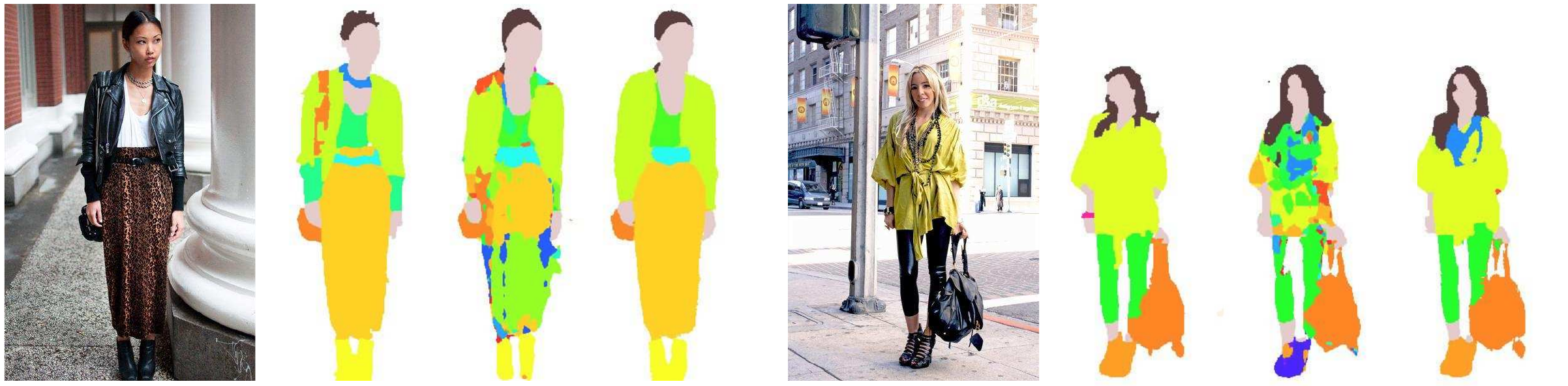}
        \\
        \makebox[\columnwidth]{
          \begin{scriptsize}
            \begin{tabular}{lllllllll}
			\fcolorbox[RGB]{0,0,0}{255,58,0}{} blazer & \fcolorbox[RGB]{0,0,0}{255,115,0}{} bag & \fcolorbox[RGB]{0,0,0}{255,144,0}{} shoes & \fcolorbox[RGB]{0,0,0}{255,202,0}{} skirt & \fcolorbox[RGB]{0,0,0}{250,255,0}{} boots & \fcolorbox[RGB]{0,0,0}{221,255,0}{} blouse & \\\fcolorbox[RGB]{0,0,0}{192,255,0}{} jacket & \fcolorbox[RGB]{0,0,0}{48,255,0}{} shirt & \fcolorbox[RGB]{0,0,0}{0,255,10}{} leggings & \fcolorbox[RGB]{0,0,0}{0,255,96}{} top & \fcolorbox[RGB]{0,0,0}{0,255,125}{} cardigan & \fcolorbox[RGB]{0,0,0}{0,255,241}{} belt & \\\fcolorbox[RGB]{0,0,0}{0,125,255}{} necklace & \fcolorbox[RGB]{0,0,0}{255,0,144}{} watch & \fcolorbox[RGB]{0,0,0}{64,32,32}{} hair & \fcolorbox[RGB]{0,0,0}{226,196,196}{} skin & \\

            \end{tabular}
          \end{scriptsize}
        }
        \caption{Fashionista v0.2}
    \end{subfigure}
    \\

    \begin{subfigure}{\columnwidth}
        \includegraphics[width=\columnwidth]{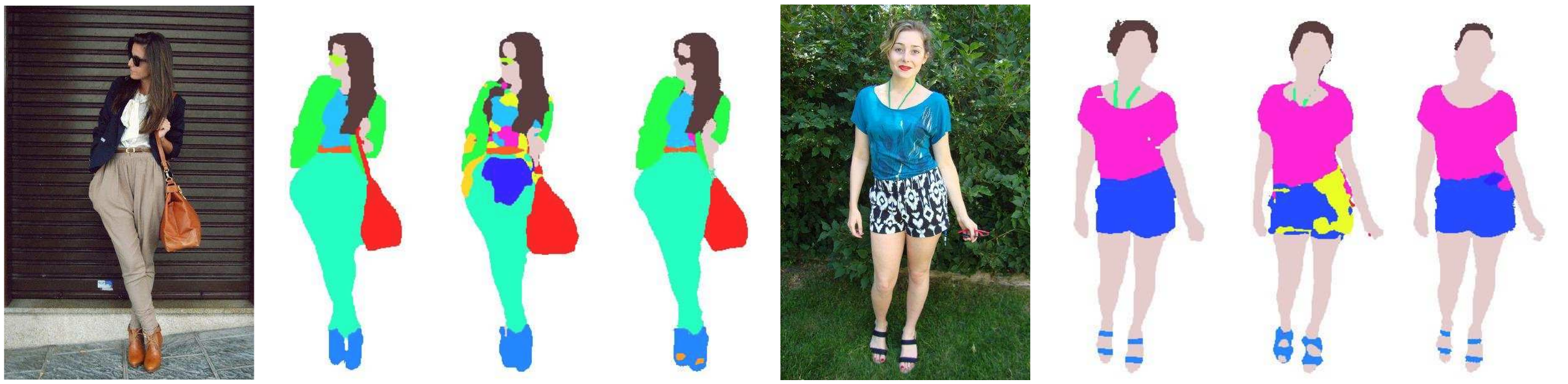}
        \\
        \makebox[\textwidth]{
        \begin{scriptsize}
          \begin{tabular}{lllllllll}
\fcolorbox[RGB]{0,0,0}{226,196,196}{} skin & \fcolorbox[RGB]{0,0,0}{64,32,32}{} hair & \fcolorbox[RGB]{0,0,0}{255,0,0}{} bag & \fcolorbox[RGB]{0,0,0}{255,70,0}{} belt & \fcolorbox[RGB]{0,0,0}{255,139,0}{} boots & \fcolorbox[RGB]{0,0,0}{255,209,0}{} coat & \\\fcolorbox[RGB]{0,0,0}{232,255,0}{} dress & \fcolorbox[RGB]{0,0,0}{162,255,0}{} glasses & \fcolorbox[RGB]{0,0,0}{0,255,46}{} jacket & \fcolorbox[RGB]{0,0,0}{0,255,116}{} necklace & \fcolorbox[RGB]{0,0,0}{0,255,185}{} pants/jeans & \fcolorbox[RGB]{0,0,0}{0,185,255}{} blouse & \\\fcolorbox[RGB]{0,0,0}{0,116,255}{} shoes & \fcolorbox[RGB]{0,0,0}{0,46,255}{} shorts & \fcolorbox[RGB]{0,0,0}{23,0,255}{} skirt & \fcolorbox[RGB]{0,0,0}{255,0,209}{} top/t-shirt & \fcolorbox[RGB]{0,0,0}{255,0,139}{} vest & \fcolorbox[RGB]{0,0,0}{255,0,70}{} bracelet & \\\\
          \end{tabular}
        \end{scriptsize}
      }
        \caption{Refined Fashionista}
    \end{subfigure}
    \\

    \begin{subfigure}{\columnwidth}
        \includegraphics[width=\columnwidth]{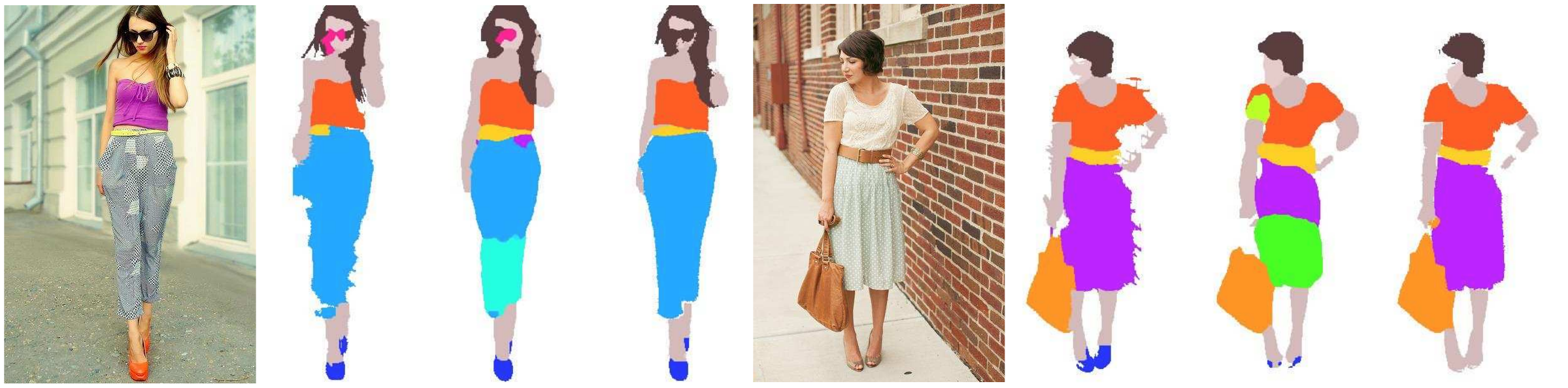}
        \\
        \makebox[\textwidth]{
        \begin{scriptsize}
          \begin{tabular}{lllllllll}
\fcolorbox[RGB]{0,0,0}{255,67,0}{} T-shirt & \fcolorbox[RGB]{0,0,0}{255,133,0}{} bag & \fcolorbox[RGB]{0,0,0}{255,200,0}{} belt & \fcolorbox[RGB]{0,0,0}{177,255,0}{} blouse & \fcolorbox[RGB]{0,0,0}{44,255,0}{} dress & \fcolorbox[RGB]{0,0,0}{226,196,196}{} face & \\\fcolorbox[RGB]{0,0,0}{64,32,32}{} hair & \fcolorbox[RGB]{0,0,0}{0,255,222}{} jeans & \fcolorbox[RGB]{0,0,0}{0,155,255}{} pants & \fcolorbox[RGB]{0,0,0}{0,22,255}{} shoe & \fcolorbox[RGB]{0,0,0}{206,176,176}{} skin & \fcolorbox[RGB]{0,0,0}{177,0,255}{} skirt & \\\fcolorbox[RGB]{0,0,0}{255,0,133}{} sunglass & \\

          \end{tabular}
        \end{scriptsize}
      }
        \caption{CFPD}
    \end{subfigure}

    \caption{Successful cases in (a) Fashionista v0.2, (b) Refined Fashionista, and (c) CFPD. The figure shows an input image, a ground truth, the output of FCN-8s and our outfit filtering with CRF from left to right respectively.}
    \label{res_good}
\end{figure}

\begin{figure}[t]
    \centering
    
    \begin{subfigure}{\columnwidth}
        \includegraphics[width=\columnwidth]{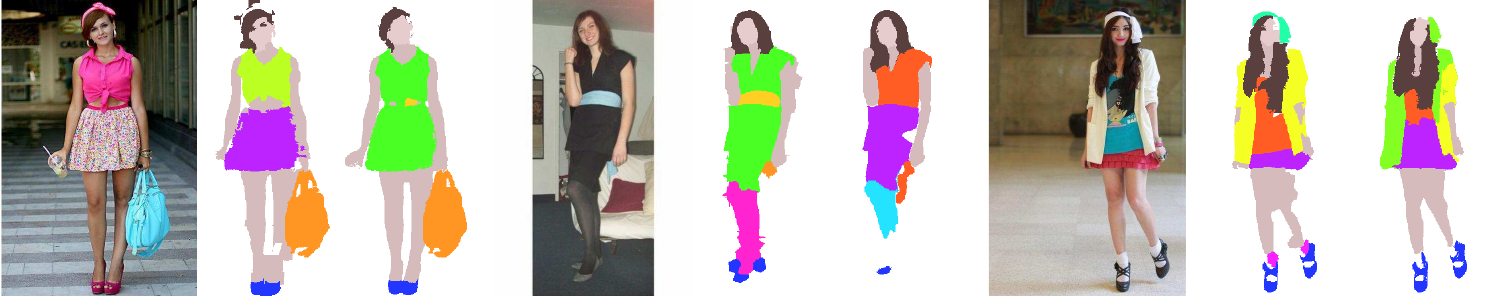}
        \caption{Prediction errors}
    \end{subfigure}
    \begin{subfigure}{\columnwidth}
        \includegraphics[width=\columnwidth]{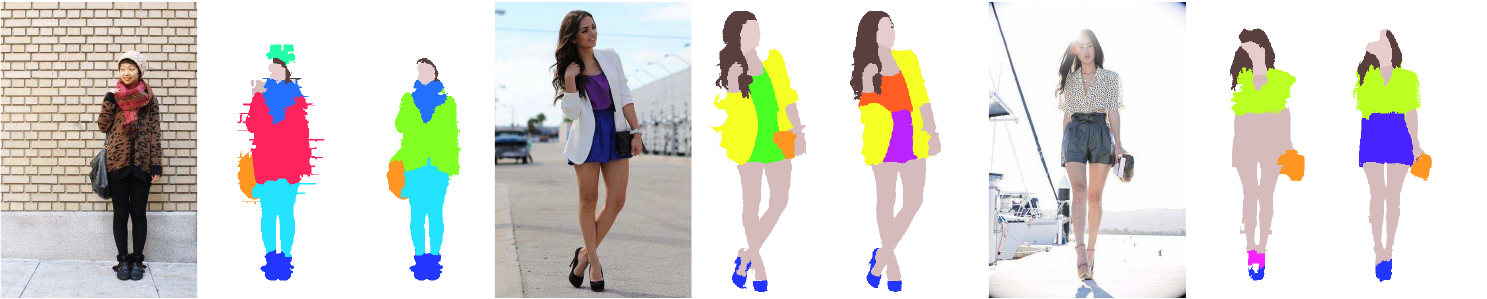}
        \caption{Dataset problem}
    \end{subfigure}
    \begin{subfigure}{\columnwidth}
        \makebox[\columnwidth]{
          \begin{scriptsize}
              \begin{tabular}{lllllllll}

          \fcolorbox[RGB]{0,0,0}{255,67,0}{} t-shirt & \fcolorbox[RGB]{0,0,0}{255,133,0}{} bag & \fcolorbox[RGB]{0,0,0}{255,200,0}{} belt & \fcolorbox[RGB]{0,0,0}{244,255,0}{} blazer & \fcolorbox[RGB]{0,0,0}{177,255,0}{} blouse & \fcolorbox[RGB]{0,0,0}{111,255,0}{} coat &\\ \fcolorbox[RGB]{0,0,0}{44,255,0}{} dress & \fcolorbox[RGB]{0,0,0}{226,196,196}{} face & \fcolorbox[RGB]{0,0,0}{64,32,32}{} hair & \fcolorbox[RGB]{0,0,0}{0,255,155}{} hat & \fcolorbox[RGB]{0,0,0}{0,222,255}{} legging & \fcolorbox[RGB]{0,0,0}{0,89,255}{} scarf &\\ \fcolorbox[RGB]{0,0,0}{0,22,255}{} shoe & \fcolorbox[RGB]{0,0,0}{44,0,255}{} shorts & \fcolorbox[RGB]{0,0,0}{206,176,176}{} skin & \fcolorbox[RGB]{0,0,0}{177,0,255}{} skirt & \fcolorbox[RGB]{0,0,0}{244,0,255}{} socks & \fcolorbox[RGB]{0,0,0}{255,0,200}{} stocking &\\ \fcolorbox[RGB]{0,0,0}{255,0,67}{} sweater & \\
              \end{tabular}
          \end{scriptsize}
        }
    \end{subfigure}

    \caption{Failure cases in CFPD. Each triple shows an input image, a ground truth, and the output of our outfit filtering with CRF respectively. Failure is either caused by (a) the model (prediction error), or (b) the dataset (clothing ambiguity or incorrect annotation).}
    \label{fig:res_bad}
\end{figure}

Figure \ref{res_good} shows successful parsing results in Fashionista v0.2, Refined Fashionista, and CFPD, using the baseline FCN-8s and our outfit filtering with CRF. Our model can produce pixel-wise segmentation that are sometimes more precise than ground-truth annotations based on superpixels. In some case, our model can correctly identify small items that were missing in the ground truth, such as {\it necklace} in the right half of Figure \ref{res_good}a, even though they were counted as mistakes in the benchmark.

Figure \ref{fig:res_bad} lists some of the failure cases from CFPD dataset. Failures can happen either because of the model or the dataset. The common error happening in the model-side is the confusion between clothing, such as making a mistake on {\it dress} vs. {\it top} and {\it skirt} combination. Such confusion can happen together in the same image, and produces a mixture of incompatible segments in the foreground region (right in figure \ref{fig:res_bad}a). We observe the common confusion among: outers (jacket, blazer, coat, sweater), inners (t-shirt, tops, shirt, blouse, dress), long-sleeves (tops, sweater), bottoms (jeans, pants, leggings), leggings (leggings, stockings, tights, long socks), or (boots, high sneakers). Some of the items are even difficult for humans to distinguish depending on the visibility. Our outfit filtering with CRF disambiguate this clothing-specific confusion and improves the segmentation quality, but some prediction errors still remain.


There is a noticeable error due to the annotation quality. The bottom row of Figure \ref{fig:res_bad} depicts cases when the ground-truth annotation is incorrect, even if our model can predict appropriate labels. The major reason of the annotation error is from 1) the inability of superpixel algorithms to respect object boundaries (bottom-center and right) and 2) human mistakes between the ambiguous categories (bottom-left). Such annotation issue introduces unreasonable artifact in the existing benchmarks.


\subsection{In-depth analysis}


\subsubsection{How well outfit prediction performs?}
Our outfit encoder learns to predict the set of applicable clothing categories as binary classification. Here, we report the performance of this prediction after the side-path encoder training. Table \ref{attr_avg_performance} summarizes the results in terms of average accuracy, precision, recall, and F1, for three datasets. Note that there are categories that always exist in the image (e.g., skin or background), but we do not exclude such labels in the evaluation.

\begin{table}[t]
  \centering
  \footnotesize
  \caption{Average performance of outfit prediction [\%].}
  \label{attr_avg_performance}
  \begin{tabular}{lcccc}
  \hline
  Dataset             & Accuracy & Precision & Recall & F1 \\ \hline
  Fashionista v0.2    & 88.66 & 25.21 & 26.57 & 25.46 \\
  Refined Fashionista & 73.36 & 39.29 & 46.06 & 40.00 \\
  CFPD                & 88.57 & 66.70 & 68.06 & 65.50 \\
  \hline
  \end{tabular}
\end{table}

The average accuracy in Fashionista v0.2 is 88.66\% while in Refined Fashionista the average accuracy is 73.36\%, despite the better performance in all other metrics. This counter-intuitive result comes from the fact that Fashionista v0.2 has much larger number of less frequent labels, and thus the large number of true negatives contributing to accuracy, because all other metrics ignore true negatives.


CFPD results have much better accuracy, precision, recall and F1 than both versions of Fashionista datasets. We suspect that the smaller number of images (685) in Fashionista compared to CFPD (2,682) is somehow causing overfitting and makes the overall performance lower in Fashionista than in CFPD. 

The drawback of our side-path architecture is the increased risk of overfitting against small datasets, because our outfit encoder must be trained to predict an image-level category and thus the available training size is restricted to the number of images but not pixels. However, we are able to mitigate overfitting by fine-tuning the entire network towards the segmentation loss at the end (Sec \ref{sec:training}). Using external weakly-annotated data~\cite{Yamaguchi2015,Liu2015} to learn the side-path is an alternative option. 

\subsubsection{How much overfitting happens in segmentation?} \label{sec:overfitting}
\begin{table}[t]
\centering
  \footnotesize
\caption{Training and testing segmentation performance of outfit encoder [\%IoU].}
\label{res_overfitting}
\begin{tabular}{lcc}

\hline
Dataset & Training & Testing \\ \hline
Refined Fashionista & 84.72 & 46.40  \\
CFPD & 76.95 & 51.42  \\
\hline
\end{tabular}
\end{table}


We evaluate how overfitting is happening in the final segmentation by comparing the training and testing performance. Table \ref{res_overfitting} summarizes the IoU performance of our outfit filtering without CRF on the training and testing splits in Refined Fashionista and CFPD. There is clearly a discrepancy between the performance in training and testing splits in both datasets, and the gap is more significant in Fashionista dataset. Indeed, we observed the training and testing discrepancy in all of the models including baseline FCNs. Our model achieves the state-of-the-art, but the result also suggests that we need a larger benchmark to properly evaluate clothing parsing.

\begin{figure}
	\centering
    \begin{subfigure}{0.9\columnwidth}
        \centering
        \includegraphics[width=\columnwidth]{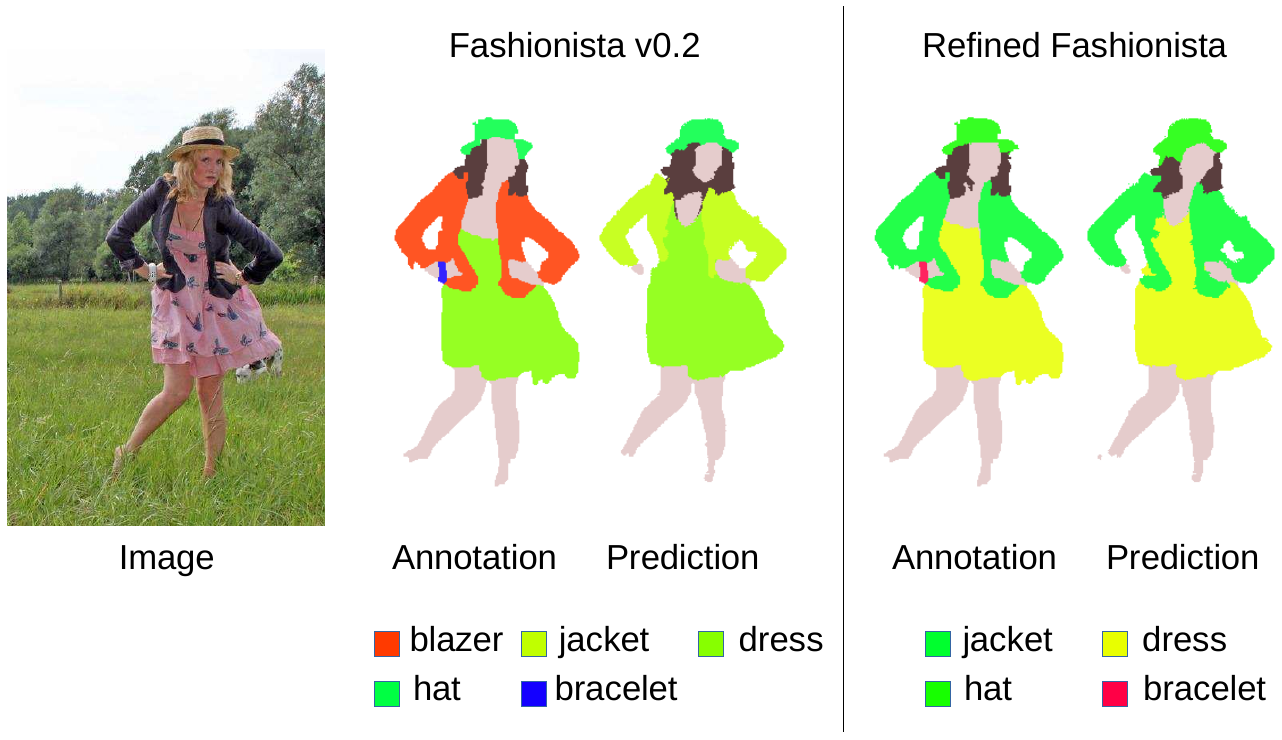}
    \end{subfigure}

    \caption{Comparison of annotation and prediction.}
    \label{fig:cmp_f02_f10}
\end{figure}

\subsubsection{Qualitative effect of refining annotation}

Figure \ref{fig:cmp_f02_f10} shows an illustrative example of how our refinement to Fashionista dataset disambiguates the segmentation. The Refined Fashionista merges some ambiguous labels, as shown in the figure that the confusion between {\it blazer} and {\it jacket} in Fashionista dataset disappeared in the refined dataset. The category-ambiguity is an inherent problem in clothing recognition, and is perhaps impossible to completely resolve. One approach might be to re-formulate the problem to allow multiple labels to be assigned to each pixel instead of the current exclusive label assignment, though that might require more annotation efforts.
 
\section{Application: outfit retrieval} \label{sec:img_retrieval}
In this section, we demonstrate the application of the outfit encoder to image retrieval. The internal representation in the outfit encoder encompasses the gist of combinatorial clothing preference. This compact representation makes an ideal use for retrieving images of certain clothing combination (outfit). We have a preliminary study of how our encoding performs in the retrieval scenario. In this study, we do not make a quantitative evaluation and instead make a small qualitative analysis, due to the challenge in defining the ground-truth in fashion similarity~\cite{7045985}.

Using Fashionista dataset, we first split the data into query and retrieval sets, and extract 256 dimensional representation from the fully-connected layer in the outfit encoder. As a baseline, we also compute the generic image feature from fc7 layer in the pre-trained VGG16 network. Our retrieval is based on Euclidean distance.

\begin{figure}[t]
    \centering
    \begin{subfigure}{\columnwidth}
        \includegraphics[width=\columnwidth]{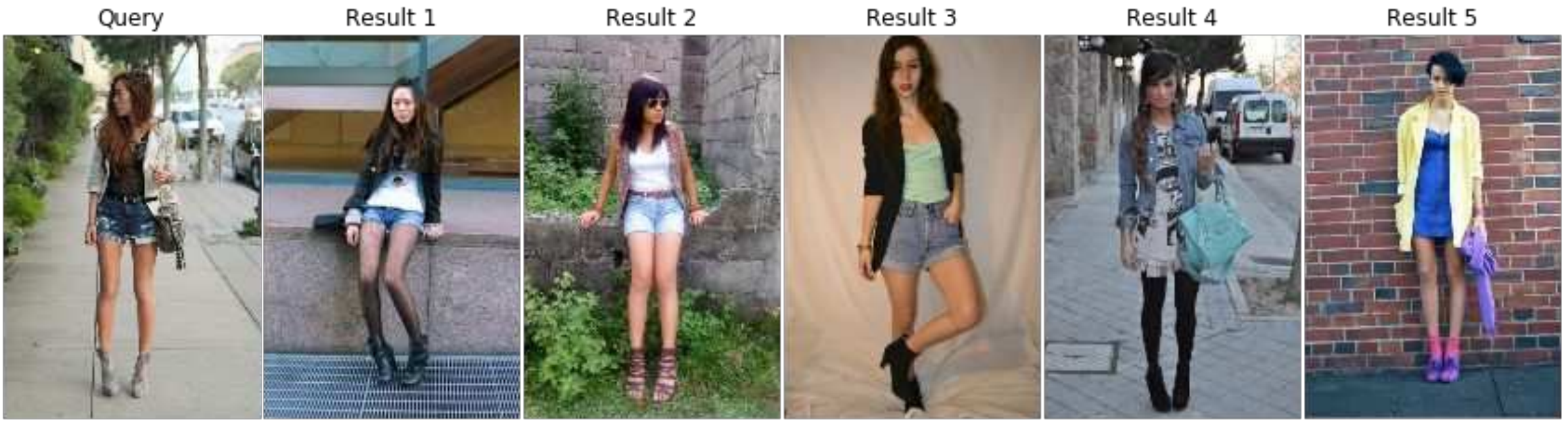}
        \caption{Using outfit encoding}
    \end{subfigure}
    \begin{subfigure}{\columnwidth}
        \includegraphics[width=\columnwidth]{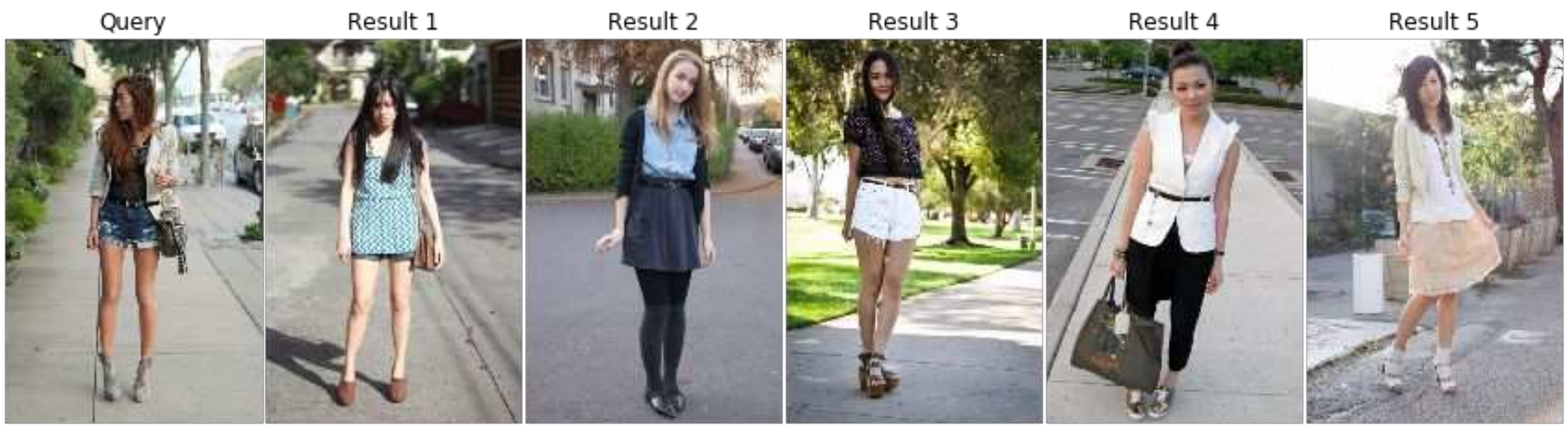}
        \caption{Using generic fc7 feature from VGG16}
    \end{subfigure}
    \caption{Image retrieval on Refined Fashionista dataset using (a) our outfit encoding, and (b) generic features from VGG16. Notice our encoding retrieves consistent clothing combination (jacket + top + shorts), while the generic feature pays attention to the background (road).}
    \label{fig:img_ret}
\end{figure}

Figure \ref{fig:img_ret} shows an example of 5 closest images retrieved using our encoder representation and the baseline fc7 feature. 
In the figure, images retrieved using our method contain the same set of garments as in query, such as jacket, top, or shorts, without concerning colors.  Retrieved images from the baseline feature seem to pay more attention on color, and the results contain the same gray and green background, but the dress-up style look rather random. We observed similar trend in other images.

The result indicates that our encoder representation clearly captures the combinatorial semantics of garments. The learned gist representation would benefit in fashion-focused applications such as outfit search or recommendation. We emphasize that the encoder does not require extra annotation for training, and the image representation comes for free in the training of the segmentation pipeline.

\section{Conclusion and future work} \label{conclusion}
This paper proposed an extension to FCN architecture to solve the clothing parsing problem. The extension includes the side-path outfit encoder to predict a set of labels in the image, and CRF to produce a consistent label assignment both in terms of clothing semantics and structure within and image. Our model can learn from a pre-trained network and the existing annotated dataset without additional data. In addition, the learned image representation in the outfit encoder is useful for similar dress-up styles thanks to the internal representation that encompasses combinatorial clothing semantics. This paper also introduced an refined annotation to Fashionista dataset for better benchmarking of clothing parsing, built with a Web-based tool to create a high-resolution pixel-based annotation. The empirical study using the Fashionista and CFPD dataset shows that our method achieves state-of-the-art parsing performance.

In the future, we wish to scale up datasets for clothing recognition with our Web-based tool, as well as to further investigate an approach to better incorporate the semantics of clothing in the prediction, for example by integrating CRF into the network~\cite{crfasrnn}. 
Also, we wish to study how human pose estimation ~\cite{toshev2014deeppose,tompson2014joint} relates to clothing parsing under fully-convolutional architecture.

{\small
\bibliographystyle{ieee}
\bibliography{my}
}

\end{document}